# DDIM sampling for Generative AIBIM, a faster intelligent structural design framework

Zhili He[1], Yu-Hsing Wang[1]

[1]Department of Civil and Environmental Engineering,
The Hong Kong University of Science and Technology, HKSAR, China

**ABSTRACT:** Generative AIBIM, a successful structural design pipeline, has proven its ability to intelligently generate high-quality, diverse, and creative shear wall designs that are tailored to specific physical conditions. However, the current module of Generative AIBIM that generates designs, known as the physics-based conditional diffusion model (PCDM), necessitates 1000 iterations for each generation due to its reliance on the denoising diffusion probabilistic model (DDPM) sampling process. This leads to a time-consuming and computationally demanding generation process. To address this issue, this study introduces the denoising diffusion implicit model (DDIM), an accelerated generation method that replaces the DDPM sampling process in PCDM. While the original DDIM was designed for DDPM and the optimization process of PCDM differs from that of DDPM, this paper designs "DDIM sampling for PCDM," which modifies the original DDIM formulations to adapt to the optimization process of PCDM. Experimental results demonstrate that DDIM sampling for PCDM can accelerate the generation process of the original PCDM by a factor of 100 while maintaining the same visual quality in the generated results. This study effectively showcases the effectiveness of DDIM sampling for PCDM in expediting intelligent structural design. Furthermore, this paper reorganizes the contents of DDIM, focusing on the practical usage of DDIM. This change is particularly meaningful for researchers who may not possess a strong background in machine learning theory but are interested in utilizing the tool effectively.

## 1 INTRODUCTION

Within the entire lifecycle of building projects, the structural design of buildings stands as a paramount and indispensable task. The accelerated pace of global urbanization has resulted in an ever-increasing demand for efficient and effective structural designs [1]. Presently, these designs are exclusively carried out by skilled structural engineers. However, this manual approach exhibits two distinct characteristics. Firstly, it heavily relies on vast structural knowledge and design expertise. Secondly, structural designs necessitate multiple iterations for refinement. Consequently, the process of structural design is time-consuming [2], labour-intensive, and inefficient [3], posing challenges in meeting the current demands [4]. Hence, there exists an urgent need for an innovative and intelligent design paradigm to address these challenges and alleviate the aforementioned issues faced by practitioners.

In the present era, rapidly evolving and widely followed AI technologies have ushered in revolutionary solutions across various industries [5,6]. The data-driven paradigm of AI [7] possesses two remarkable attributes: (1) AI models possess virtually limitless memory capacity, enabling them to retain and represent vast amounts of data characteristics. Notably, models such as ChatGPT and GPT-4 can effectively encapsulate the entirety of human knowledge; (2) AI models operate with automation and efficiency [8], rendering their inference time nearly negligible when compared to human counterparts. Consequently, AI emerges as a natural fit for structural design tasks, and the exploration of effective utilization of AI in designing structures has become a prominent area of research.

Given the vast diversity in forms and layouts of buildings, current research efforts in the field primarily focus on the intelligent design of high-rise residential buildings utilizing shear wall systems [1-4]. This emphasis is primarily driven by two key factors: First, high-rise residential buildings constitute a substantial proportion of the overall building landscape, with design and construction demands continuously on the rise. Second, the layout of such buildings tends to exhibit a relatively regular pattern, often characterized by rectangular rooms and walls. This inherent regularity simplifies the design process for AI systems, making it more amenable to intelligent design applications.

Initially, the studies on intelligent design utilizing AI primarily employed regression-based methods. For instance, Pizarro et al. [9,10] utilized multi-layer perceptrons (MLPs) and convolutional neural networks (CNNs) to regress shear wall structure layouts. Zhao et al. [3,11] introduced graph neural networks (GNNs) to regress the graph representation of shear wall layouts. However, regression-based methods suffer from inherent drawbacks: First, they often lack diversity in their generated outputs [1]. Second, the results obtained from regression-based models do not consistently align with human perception, leading to visibly lower quality predictions [12]. Lastly, these methods tend to produce results with fewer high-fidelity details [13]. In contrast, deep generative models (DGMs), a novel AI framework, offer a promising alternative. DGMs can generate more intricate, diverse, and perceptually plausible images when compared to regression-based models [12,13].

Consequently, researchers have recently turned their attention towards integrating DGMs into intelligent design methodologies. Among these DGMs, two notable ones are generative adversarial networks (GANs) [14] and the more recent diffusion models (DMs) [15,16]. Liao et al. [2] were the first to extend the application of GANs to intelligent shear wall design, proposing StructGAN [2] along with a series of variants such as StructGAN-PHY [4] and StructGAN-AE [17]. However, GANs exhibit two inherent limitations: (1) training instability and oscillation, and (2) failure to explore and cover the entire distribution of training data [18], often getting trapped in specific modes of data distribution [19]. These characteristics hinder the visual quality of the generated results from StructGAN and its variants. For instance, many designed walls fail to meet the fundamental design requirements of having horizontal or vertical boundaries and of having clear enough design drawings [1]. On the other hand, diffusion models (DMs), particularly the denoising diffusion probabilistic model (DDPM) [16], have garnered substantial attention in recent years due to their enhanced stability and superior generation capabilities. DMs have demonstrated remarkable performance not only in image synthesis but also in various downstream computer vision tasks such as deblurring [12], super-resolution [13], and denoising [20]. Moreover, DMs have found successful applications in influential commercial AI products like DALL·E[1] and Stable Diffusion[2] for image generation, as well as Sora[3] for video generation. Building upon these advancements, He et al. [1] propose Generative AIBIM, an automatic and intelligent structural design pipeline. In this framework, they introduce a novel physics-based conditional diffusion model (PCDM) for the intelligent generation of structural design drawings. Experimental results demonstrate that PCDM surpasses all GAN-based models, not only in quantitative evaluations but also in terms of visual quality and image details. Additionally, PCDM accommodates diverse and creative design requirements tailored to specific physical conditions, such as building heights and earthquake intensities.

---

[1] https://openai.com/index/dall-e-3/

[2] https://stability.ai/

[3] https://openai.com/index/sora/

Nevertheless, it is crucial to acknowledge a significant drawback of DDPM: its sampling process demands a substantial number of iterations to generate high-quality samples [21]. Since PCDM's sampling process is built upon that of DDPM, it inherits the same limitation. In their work [1], He et al. mention that PCDM requires 1000 iterations to complete a single generation, which is exceedingly time-consuming and resource-intensive. This becomes challenging when there is a need to generate multiple structural designs. To address this critical flaw in DDPM, Song et al. [21] propose the denoising diffusion implicit model (DDIM). DDIM effectively reduces the length of sampling chains while ensuring the generation of high-quality results. In fact, Song et al. [21] demonstrate that DDIM can accelerate the generation process of DDPM by a remarkable factor of 10 to 100 times.

Consequently, this study incorporates DDIM into Generative AIBIM to explore its potential for expediting the generation of structural design drawings. First, the original DDIM paper [21] primarily focuses on theoretical coherence, which may prove challenging for researchers without a strong background in machine learning and probability theory. As civil engineers, our primary concern lies in the practical utilization of the tool rather than delving into intricate mathematical theories. Therefore, this paper reorganizes the contents of DDIM, emphasizing its practical application and omitting the more obscure theoretical aspects, as presented in Subsection 2.2. Second, the original DDIM formulation is specifically designed for DDPM and cannot be directly applied to PCDM due to the differing optimization processes involved. Consequently, this study modifies the formulations to ensure their integration with PCDM, resulting in a modified version of DDIM known as "DDIM sampling for PCDM," as elucidated in Subsection 2.3. The experimental results, presented in Section 3, demonstrate that DDIM can accelerate the generation process of PCDM by a factor of 100 while maintaining the same level of generation quality. DDIM sampling for PCDM has been integrated into the code of Generative AIBIM, which is publicly available at https://github.com/hzlbbfrog/Generative-BIM.

## 2 METHOD
### 2.1 Generative AIBIM
In their seminal work [1], He et al. integrate Building Information Modeling (BIM) with the intelligent design of shear walls, presenting an innovative Generative AIBIM pipeline for structural design. This pipeline not only expands the application range of BIM but also complements the existing methodologies for intelligent design that are confined to CAD drawings. The Generative AIBIM framework encompasses four stages: (1) Stage I: from BIM models to architectural design drawings; (2) Stage II: intelligent structural design based on generative AI; (3) Stage III: from structural design drawings to BIM models; and (4) Stage IV: BIM-integrated applications. It is worth noting that Stage II stands as the pivotal component of this framework. In this stage, He et al. [1] devise a two-stage generation framework, drawing inspiration from the human drawing process. Their algorithm initially generates line drawings based on input canvases in Stage 1. In Stage 2, the line drawings are subsequently colored into structural design drawings (refer to Fig. 3 in [1] for enhanced comprehension). In the two-stage generation framework, Stage 1 is the core part, where He et al. [1] propose PCDM for generating line drawings based on DDPM. PCDM includes three distinct processes: the forward diffusion process, the reverse denoising process (also known as the sampling process), and the optimization process. These processes are summarized as follows:

In the forward diffusion process, PCDM injects Gaussian noise $\epsilon \in \mathbb{R}^D$ step-by-step through a $T$-step Markov chain:

$$q(\mathbf{x}_t|\mathbf{x}_{t-1}) \triangleq \mathcal{N}(\mathbf{x}_t; \sqrt{1-\beta_t}\mathbf{x}_{t-1}, \beta_t \mathbf{I}_D), \tag{1}$$

where $D$ denotes the dimension of a line drawing, $t \in \{1,2,\dots,T\}$ is the time step, $\mathbf{x}_t$ means the noise sample, and clearly, $\mathbf{x}_0$ is the clean sample, namely, a canvas. $\mathbf{I}_D$ denotes the identity matrix. $\beta_t$ is the noise schedule and is defined by the following formulation:

$$\beta_t = 1 - \frac{\bar{\alpha}_t}{\bar{\alpha}_{t-1}}, \tag{2}$$

where $\bar{\alpha}_t$ is a cosine function and $\beta_t \in (0,1)$ for $\forall t$. Furthermore, we can obtain $\mathbf{x}_t$ through a single sampling for arbitrary $t$:

$$\mathbf{x}_t \sim q(\mathbf{x}_t|\mathbf{x}_0) = \mathcal{N}(\mathbf{x}_t; \sqrt{\bar{\alpha}_t}\mathbf{x}_0, (1-\bar{\alpha}_t)\mathbf{I}_D). \tag{3}$$

By the reparameterization trick, we can derive the formulation of $\mathbf{x}_t$:

$$\mathbf{x}_t = \sqrt{\bar{\alpha}_t}\mathbf{x}_0 + \sqrt{1-\bar{\alpha}_t}\epsilon. \tag{4}$$

where $\epsilon \sim \mathcal{N}(0, \mathbf{I}_D)$.

In the reverse denoising process, PCDM wants to regenerate clean samples $\hat{\mathbf{x}}_0$ from $\mathbf{x}_T$ and to make $\hat{\mathbf{x}}_0$ follow the real data distribution. First, we need to determine $\mathbf{x}_T$. Based on Eq. (3), clearly, when $T$ is a large number, approximately, $\mathbf{x}_T \sim \mathcal{N}(0, \mathbf{I}_D)$. Thus, we can start from $\hat{\mathbf{x}}_T \sim \mathcal{N}(0, \mathbf{I}_D)$ and the difference between the distributions of $\mathbf{x}_T$ and $\hat{\mathbf{x}}_T$ can be considered negligible. In PCDM, $T$ is set to 1000. Further, we can formulate $q(\mathbf{x}_{t-1}|\mathbf{x}_t, \mathbf{x}_0)$:

$$q(\mathbf{x}_{t-1}|\mathbf{x}_t, \mathbf{x}_0) = \mathcal{N}\left(\mathbf{x}_{t-1}; \tilde{\boldsymbol{\mu}}_t(\mathbf{x}_t, \mathbf{x}_0), \widetilde{\boldsymbol{\Sigma}}_t\right), \tag{5}$$

where

$$\tilde{\boldsymbol{\mu}}_t(\mathbf{x}_t, \mathbf{x}_0) = \frac{\sqrt{1-\beta_t}(1-\bar{\alpha}_{t-1})}{1-\bar{\alpha}_t}\mathbf{x}_t + \frac{\sqrt{\bar{\alpha}_{t-1}}\beta_t}{1-\bar{\alpha}_t}\mathbf{x}_0, \tag{6}$$

and

$$\widetilde{\boldsymbol{\Sigma}}_t = \tilde{\beta}_t \mathbf{I}_D, \tilde{\beta}_t = \frac{1-\bar{\alpha}_{t-1}}{1-\bar{\alpha}_t}\beta_t. \tag{7}$$

In the reverse denoising process, PCDM uses a transition distribution $p_{\boldsymbol{\theta}}(\mathbf{x}_{t-1}|\mathbf{x}_t, \mathbf{y}, d)$ to substitute $q(\mathbf{x}_{t-1}|\mathbf{x}_t, \mathbf{x}_0)$ and follows DDPM to run a reverse Markov chain to sample data from $t = T$ to $t = 1$ to obtain the ultimate generation result $\hat{\mathbf{x}}_0$ based on the distribution $p_{\boldsymbol{\theta}}(\mathbf{x}_{t-1}|\mathbf{x}_t, \mathbf{y}, d)$. The schematic diagrams of the forward and reverse processes are shown in Fig. 1. Clearly, PCDM needs $T = 1000$ iterations to finish one generation.

The goal of the optimization process is to efficiently model $p_{\boldsymbol{\theta}}(\mathbf{x}_{t-1}|\mathbf{x}_t, \mathbf{y}, d)$ to make it approximate $q(\mathbf{x}_{t-1}|\mathbf{x}_t, \mathbf{x}_0)$. First, $p_{\boldsymbol{\theta}}(\mathbf{x}_{t-1}|\mathbf{x}_t, \mathbf{y}, d)$ is also defined as a Gaussian distribution:

$$p_{\boldsymbol{\theta}}(\mathbf{x}_{t-1}|\mathbf{x}_t, \mathbf{y}, d) \triangleq \mathcal{N}(\mathbf{x}_{t-1}; \boldsymbol{\mu}_{\boldsymbol{\theta}}(\mathbf{x}_t, t, \mathbf{y}, d), \boldsymbol{\Sigma}_{\boldsymbol{\theta}}(\mathbf{x}_t, t, \mathbf{y}, d)), \tag{8}$$

For simplicity, $\boldsymbol{\Sigma_\theta}$ is set to $\widetilde{\boldsymbol{\Sigma}}_t$. Then, the question is simplified into "how to model $\boldsymbol{\mu_\theta}$ to make it close to $\tilde{\boldsymbol{\mu}}_t$. DDPM first combines Eq. (4) and Eq. (6) to eliminate $\mathbf{x}_0$:

$$\tilde{\boldsymbol{\mu}}_t = \frac{1}{\sqrt{1-\beta_t}}\left(\mathbf{x}_t - \frac{\beta_t}{\sqrt{1-\bar{\alpha}_t}}\boldsymbol{\epsilon}\right). \tag{9}$$

Then, it builds a neural network $\hat{\boldsymbol{\epsilon}} = \boldsymbol{\epsilon}_\theta(\mathbf{x}_t, t, \mathbf{y}, d)$ to approximate $\boldsymbol{\epsilon}$, and $\boldsymbol{\mu_\theta}$ can be defined as

$$\boldsymbol{\mu_\theta} = \frac{1}{\sqrt{1-\beta_t}}\left(\mathbf{x}_t - \frac{\beta_t}{\sqrt{1-\bar{\alpha}_t}}\hat{\boldsymbol{\epsilon}}\right). \tag{10}$$

PCDM proposes another solution. It first divides $\mathbf{x}_0$ into two parts: shear walls $\mathbf{s}_0$ and infill walls $\mathbf{y}$. Then, a neural network is constructed to approximate $\mathbf{s}_0$ : $\hat{\mathbf{s}}_0^t = f_\theta(\mathbf{x}_t, t, \mathbf{y}, d)$ , so, $\hat{\mathbf{x}}_0^t = f_\theta(\mathbf{x}_t, t, \mathbf{y}, d) + \mathbf{y}$. Next, plug $\hat{\mathbf{x}}_0^t$ into Eq. (6) to model $\boldsymbol{\mu_\theta}$:

$$\boldsymbol{\mu_\theta} = \frac{\sqrt{1-\beta_t}(1-\bar{\alpha}_{t-1})}{1-\bar{\alpha}_t}\mathbf{x}_t + \frac{\sqrt{\bar{\alpha}_{t-1}}\beta_t}{1-\bar{\alpha}_t}\hat{\mathbf{x}}_0^t, \tag{11}$$

As for the neural network, PCDM introduces an innovative attention block that includes a self-attention block [22] and a parallel cross-attention block, along with an adaptive Instance Normalization block to facilitate the fusion of data from different domains.

## 2.2 DDIM Sampling

As elucidated in Subsection 2.1, the original PCDM follows the sampling process of DDPM, necessitating $T$ iterations to obtain $\hat{\mathbf{x}}_0$, and $T$ is a large number to ensure that the distribution of $\mathbf{x}_T$ close to $\mathcal{N}(0, \mathbf{I}_D)$ ($T$ is set to 1000 in PCDM). Then, DDIM is proposed to expedite the generation process of DDPM by shortening sampling chains.

First of all, following the optimization process of DDPM, the prediction object of the neural network is still noise: $\hat{\boldsymbol{\epsilon}} = \boldsymbol{\epsilon}_\theta(\mathbf{x}_t, t)$. Then, the core idea of DDIM is that if we have a subset of the original time steps $\{1,2,\dots,T\}$: $\tau = \{\tau_1, \tau_2, \dots, \tau_S\}$, where $\tau_1 < \tau_2 < \cdots < \tau_S \in [1, T]$, and $S$ denotes the total time step of the sub-sequence. Clearly, $S < T$. Further, if the ultimate sample generated by the sub-sequence, $\hat{\mathbf{x}}_0$, has the same quality as the data sampled along $\{1,2,\dots,T\}$, we only need to consider the sampling along the sub-sequence, and the number of iterations is reduced to $N$ from $T$. Thus, generation efficiency is increased. DDIM proves that by modifying the sampling rule, different sub-sequences are equivalent to the original $\{1,2,\dots,T\}$ and can be adopted as the sampling chain. The implementation details of the new rule are introduced as follows:

The same as DDPM, we still start from the end of data and sample $\hat{\mathbf{x}}_{\tau_S}$ from $\mathcal{N}(0, \mathbf{I}_D)$. $\tau_S$ still should be a large number, and $\tau_S$ is set to $T$ for simplicity. Similarly, $\tau_1$ is always set to 1. Based on mathematical induction, if we can define the transformation regulations from $\mathbf{x}_{\tau_i}$ to $\mathbf{x}_{\tau_{i-1}}$, we can achieve the sampling along the new chain $\tau$. First, for different elements in $\tau$, DDIM defines a variable variance hyperparameters $\sigma$:

$$\sigma_{\tau_i}(\eta) = \eta\sqrt{\frac{1-\bar{\alpha}_{\tau_{i-1}}}{1-\bar{\alpha}_{\tau_i}}}\sqrt{1-\frac{\bar{\alpha}_{\tau_i}}{\bar{\alpha}_{\tau_{i-1}}}}, \tag{12}$$

where $i$ is the index of $\tau$, and $\eta \geq 0$ is a controllable hyperparameter. Based on the above equation, we can define the direction pointing to $\mathbf{x}_{\tau_i}$:

$$\mathbf{x}_{\tau_i}^d = \sqrt{1 - \bar{\alpha}_{\tau_{i-1}} - \sigma_{\tau_i}^2} \cdot \boldsymbol{\epsilon}_{\boldsymbol{\theta}}(\mathbf{x}_{\tau_i}, \tau_i), \tag{13}$$

where $\mathbf{x}_{\tau_i}^d$ denotes the direction pointing to $\mathbf{x}_{\tau_i}$. Recall Eq. (4), we have derived the formulation of $\mathbf{x}_t$. Since we have known $\mathbf{x}_{\tau_i}$ and $\boldsymbol{\epsilon}_{\boldsymbol{\theta}}(\mathbf{x}_{\tau_i}, \tau_i)$, we can represent $\mathbf{x}_0$ by $\mathbf{x}_{\tau_i}$ and $\boldsymbol{\epsilon}_{\boldsymbol{\theta}}(\mathbf{x}_{\tau_i}, \tau_i)$, and this representation is the prediction of $\mathbf{x}_0$ at $t = \tau_i$:

$$\hat{\mathbf{x}}_0^{\tau_i} = \frac{\mathbf{x}_{\tau_i} - \sqrt{1 - \bar{\alpha}_{\tau_i}}\boldsymbol{\epsilon}_{\boldsymbol{\theta}}(\mathbf{x}_{\tau_i}, \tau_i)}{\sqrt{\bar{\alpha}_{\tau_i}}}, \tag{14}$$

where, $\hat{\mathbf{x}}_0^{\tau_i}$ is the prediction of $\mathbf{x}_0$. Combining Eqs. (13) and (14), we can formulate $\mathbf{x}_{\tau_{i-1}}$:

$$\begin{aligned}\mathbf{x}_{\tau_{i-1}} &= \sqrt{\bar{\alpha}_{\tau_{i-1}}}\hat{\mathbf{x}}_0^{\tau_i} + \mathbf{x}_{\tau_i}^d + \sigma_{\tau_i}\boldsymbol{\epsilon}_{\tau_i} \\ &= \sqrt{\bar{\alpha}_{\tau_{i-1}}}\left(\frac{\mathbf{x}_{\tau_i} - \sqrt{1 - \bar{\alpha}_{\tau_i}}\boldsymbol{\epsilon}_{\boldsymbol{\theta}}(\mathbf{x}_{\tau_i}, \tau_i)}{\sqrt{\bar{\alpha}_{\tau_i}}}\right) + \sqrt{1 - \bar{\alpha}_{\tau_{i-1}} - \sigma_{\tau_i}^2} \cdot \boldsymbol{\epsilon}_{\boldsymbol{\theta}}(\mathbf{x}_{\tau_i}, \tau_i) + \sigma_{\tau_i}\boldsymbol{\epsilon}_{\tau_i},\end{aligned} \tag{15}$$

where $\boldsymbol{\epsilon}_{\tau_i}$ follows a standard normal distribution: $\boldsymbol{\epsilon}_{\tau_i} \sim \mathcal{N}(0, \mathbf{I})$. This is equivalent to Eq. (12) in [21]. It is important to note that in DDIM sampling, $\eta$ is set to 0, resulting in $\sigma_{\tau_i} = 0$, and DDIM sampling is transferred into DDPM sampling if $\eta$ is assigned a value of 1 [21]. Then, we can notice that the DDIM sampling process is deterministic in comparison to DDPM sampling. The above content is the new sampling rule, and we omit the theoretical part, specifically, the proof of equivalence between subsequences and the original sequence $\{1, 2, \ldots, T\}$.

### 2.3 DDIM sampling for PCDM

The original DDIM is designed for DDPM, and in the optimization process of DDPM, the prediction of the neural network is noise $\boldsymbol{\epsilon}_{\boldsymbol{\theta}}$. While in the optimization process of PCDM, the prediction is shear walls $\hat{\mathbf{s}}_0^t = f_{\boldsymbol{\theta}}(\mathbf{x}_t, t, \mathbf{y}, d)$, we need to make modifications to the original DDIM sampling formulations. As shown in Fig. 1, the modified DDIM sampling process is named DDIM sampling for PCDM, which is detailed as follows:

First, we can directly obtain the prediction of $\mathbf{x}_0$ at $t = \tau_i$, which is different from Eq. (14):

$$\hat{\mathbf{x}}_0^{\tau_i} = f_{\boldsymbol{\theta}}(\mathbf{x}_{\tau_i}, \tau_i, \mathbf{y}, d) + \mathbf{y}. \tag{16}$$

Plugging the above equation into Eq. (14), we can formulate $\boldsymbol{\epsilon}_{\boldsymbol{\theta}}$:

$$\boldsymbol{\epsilon}_{\boldsymbol{\theta}} = \frac{\mathbf{x}_{\tau_i} - \sqrt{\bar{\alpha}_{\tau_i}}\hat{\mathbf{x}}_0^{\tau_i}}{\sqrt{1 - \bar{\alpha}_{\tau_i}}} = \frac{\mathbf{x}_{\tau_i} - \sqrt{\bar{\alpha}_{\tau_i}}(f_{\boldsymbol{\theta}}(\mathbf{x}_{\tau_i}, \tau_i, \mathbf{y}, d) + \mathbf{y})}{\sqrt{1 - \bar{\alpha}_{\tau_i}}}. \tag{17}$$

The new direction pointing to $\mathbf{x}_{\tau_i}$ is

$$\mathbf{x}_{\tau_i}^d = \sqrt{1 - \bar{\alpha}_{\tau_{i-1}} - \sigma_{\tau_i}^2}\boldsymbol{\epsilon}_{\boldsymbol{\theta}} = \sqrt{1 - \bar{\alpha}_{\tau_{i-1}} - \sigma_{\tau_i}^2}\frac{\mathbf{x}_{\tau_i} - \sqrt{\bar{\alpha}_{\tau_i}}(f_{\boldsymbol{\theta}}(\mathbf{x}_{\tau_i}, \tau_i, \mathbf{y}, d) + \mathbf{y})}{\sqrt{1 - \bar{\alpha}_{\tau_i}}}. \tag{18}$$

At this time, $\mathbf{x}_{\tau_{i-1}}$ is still equal to $\sqrt{\bar{\alpha}_{\tau_{i-1}}}\hat{\mathbf{x}}_0^{\tau_i} + \mathbf{x}_{\tau_i}^d + \sigma_{\tau_i}\boldsymbol{\epsilon}_{\tau_i}$, However, $\hat{\mathbf{x}}_0^{\tau_i}$ and $\mathbf{x}_{\tau_i}^d$ should be replaced with Eqs. (16) and (17), respectively.

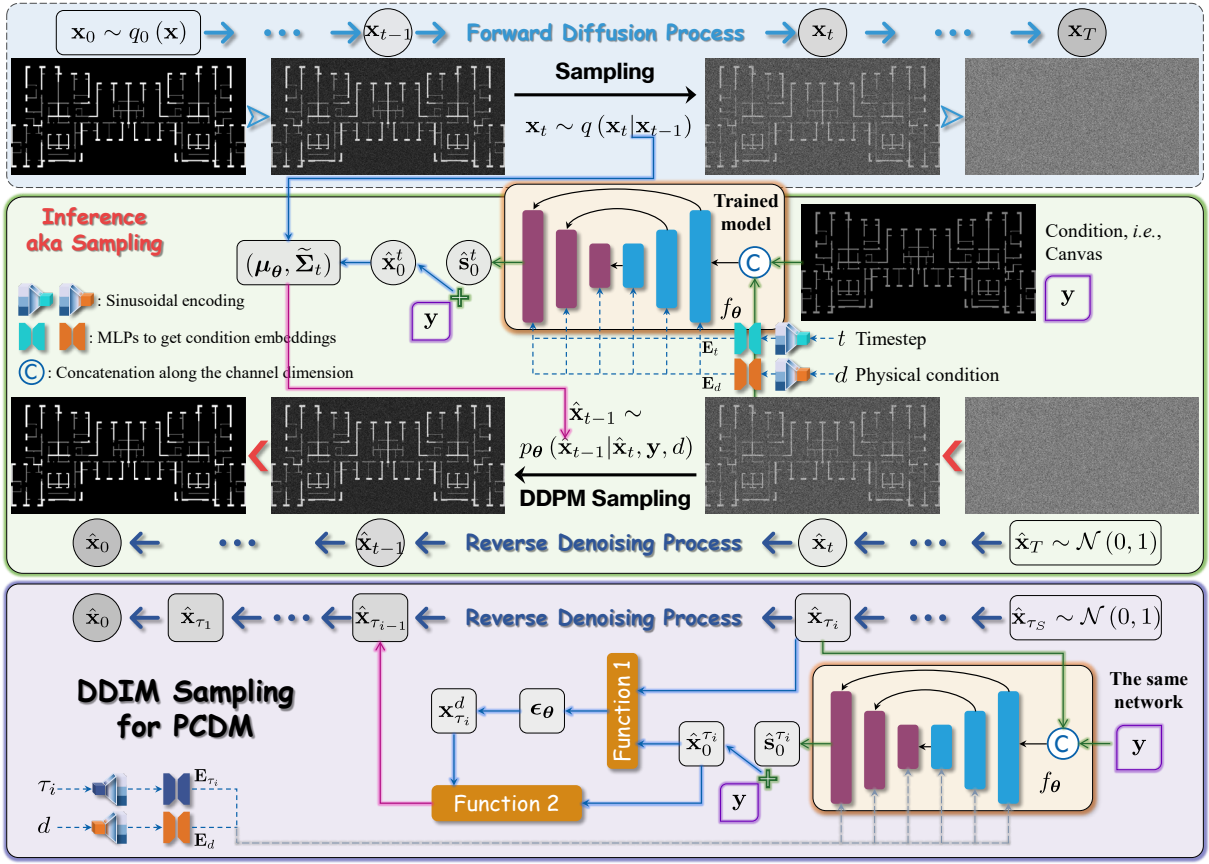

Fig. 1. Schematic diagram of the forward diffusion process and the reverse denoising processes.

## 3 EXPERIMENTS

### 3.1 Details of implementation

Dataset. The dataset used in this study is Modified-dataset proposed in [1] and can be publicly available at https://github.com/hzlbbfrog/Generative-BIM. This dataset consists of 700 training images and 24 images in the test set.

Inference environment. The hardware and software environments are the same with [1]. For example, the GPU is a Nvidia GeForce RTX 4090, and the version of PyTorch is 1.12.1.

Evaluation metrics. The Fréchet Inception Distance (FID) is adopted to evaluate the quality of the generated images following [1]. FID is consistent with human perception and is capable of quantitatively evaluating the visual quality of the generated samples.

### 3.2 Comparison of the generation results of PCDM and DDIM sampling for PCDM

Extensive experiments are conducted in this subsection to evaluate the effectiveness of DDIM sampling for PCDM, which stands as the central contribution of this study. Specifically, we employ the well-trained model in [1] and execute the reverse denoising process using both the original sampling method in PCDM (i.e., DDPM sampling) and DDIM sampling for PCDM outlined in Subsection 2.3. Additionally, to explore the generation efficiency, we consider four different lengths of sub-sequences in DDIM sampling: $S=10$, $S=20$, $S=50$, and $S=100$. Correspondingly, the generation speeds are approximately accelerated by 100×, 50×, 20×, and 10×, respectively (recall that the sequence length of PCDM is 1000). The FID results, presented in Table 1, indicate that the FID values are remarkably close, with no discernible differences. Furthermore, the qualitative comparison results between the original PCDM and the proposed DDIM sampling for PCDM are depicted in Fig. 2. Evidently, the perceptual quality and details of the generated outputs are strikingly similar, making it arduous to distinguish which method is superior. The qualitative comparison aligns with the quantitative findings, leading us to draw the conclusion that DDIM sampling for PCDM can accelerate the reverse sampling process of the original PCDM by up to 100 times while maintaining the same level of visual quality. This shows the proposed DDIM sampling for PCDM is a practical solution to expedite intelligent structural design.

Table 1. Quantitative evaluation of the results obtained from different lengths of sampling sequences.

| Length of sequence | 10 | 20 | 50 | 100 | 1000 (i.e., PCDM) |
|---|---|---|---|---|---|
| FID ↓ | 15.03 | 14.22 | 14.90 | 14.39 | 14.94 |

Notes: (1) ↓ means for FID, the lower the value, the better. (2) Unlike [1], FIDs here are measured on the entire test set, namely, the combination of the three sub-datasets.

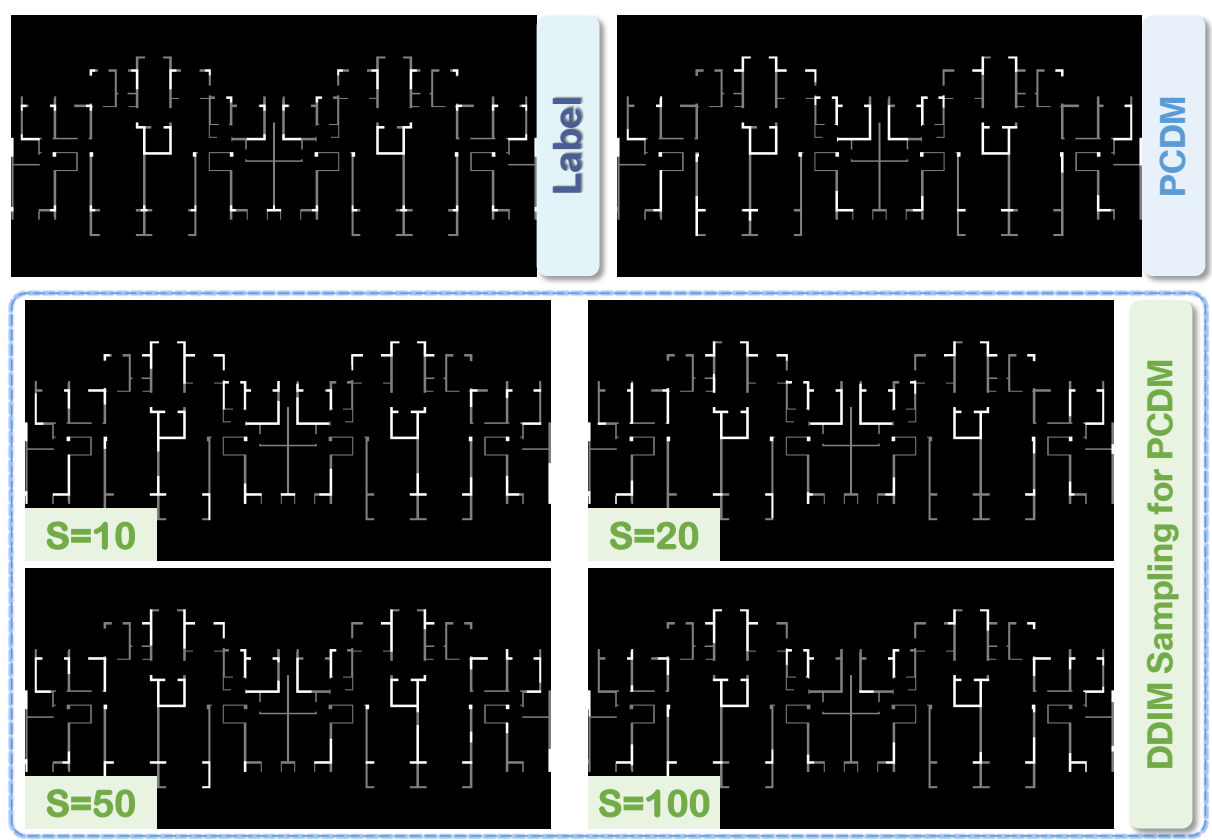

Fig. 2. Comparison of visualization results of labeling and the ones obtained from the original sampling method in PCDM and the proposed DDIM sampling for PCDM.

**4 CONCLUSIONS**

Generative AIBIM is a successful and intelligent framework for structural design, with its core module being PCDM, a novel physics-based conditional diffusion model capable of generating high-quality shear wall designs that meet diverse physical conditions. However, PCDM needs 1000 iterations to generate each high-quality sample, hindered by the nature of the adopted DDPM sampling method, which is so time-consuming and requires substantial computing resources. To address this limitation, this study introduces DDIM into the Generative AIBIM framework to expedite the generation process. The key innovations of this research are as follows: First, this paper reorganizes the content of the original DDIM paper, emphasizing its practical usage and making it more accessible to researchers who may not possess an extensive background in machine learning. This change is particularly valuable for those who are primarily interested in utilizing the tool rather than delving into its theoretical intricacies. Second, this study presents DDIM sampling for PCDM, which is a modification of the original DDIM to suit its integration into PCDM. This adaptation is necessary as the original DDIM cannot be directly embedded into PCDM. Third, extensive experiments are conducted to validate the effectiveness of DDIM sampling for PCDM. The experimental results demonstrate that DDIM sampling for PCDM can accelerate the sampling process of PCDM by 100× while maintaining the same level of visual quality in the generated outputs. Overall, this study offers a successful implementation of DDIM within the Generative AIBIM pipeline and provides compelling evidence for the efficiency of DDIM sampling for PCDM in accelerating intelligent structural design.

**ACKNOWLEDGEMENTS**
The research presented was financially supported by the Innovation Technology Fund, Midstream Research Programme for Universities [project no. MRP/003/21X], and the Hong Kong Research Grants Council [project no. 16205021]. Contributions by the anonymous reviewers are also highly appreciated.